**Universidad Tecnológica do Uruguay, Rivera, Uruguay**

# DRONES E INTELIGENCIA ARTIFICIAL PARA INVESTIGACIÓN Y COMPETICIÓN

Saravia, Victoria, victoria.saravia@estudiantes.utec.edu.uy[1]
Moraes, William, william.moraes@estudiantes.utec.edu.uy[2]

Bedin Grando, Ricardo, ricardo.bedin@utec.edu.uy[3]
Da Silva Kelbouscas, André , andre.dasilva@utec.edu.uy[4]

[1]Universidad Tecnológica del Uruguay, Rivera, Uruguay
[2]Universidad Tecnológica del Uruguay, Rivera, Uruguay

[3]Universidad Tecnológica del Uruguay, Rivera, Uruguay
[4]Universidad Tecnológica del Uruguay, Rivera, Uruguay

**ÍNDICE:**

**RESUMEN**

El área de robótica industrial está en crecimiento. Llevar a cabo tareas donde es necesaria gran precisión o procesos repetitivos ya no se dejan en manos del humano. Por consecuencia, la productividad dentro de una línea de producción aumenta. En este contexto, es considerado un área de interés siendo así un área en la que la innovación es muy importante..

Considerando esta realidad, este trabajo se enfoca en los drones o UAVs (Vehículos Aéreos no Tripulados) para su uso en la industria en general. Estos vehículos tienen un gran número de utilidades y potencial en la industria, como una herramienta de auxilio para la ingeniería civil, medicina, minería, entre otras. Sin embargo, este vehículo es limitado para su uso en locales cerrados debido a la necesidad de GPS ya que no funciona en locales cerrados. De esta forma, este trabajo presenta un UAV que funciona sin GPS, pudiendo así ser usado en espacios cerrados por ejemplo y tener buena precisión. El trabajo parte de un abordaje que utiliza visión por computadora y GPS.

***Palabras clave:*** *inteligencia artificial, dron, robótica.*

**ABSTRACT**

The area of industrial robotics is growing. Carrying out tasks where high precision or repetitive processes are required is no longer left to the human. Consequently, productivity within a production line increases. In this context, it is considered an area of interest, thus being an area in which innovation is very important.

Considering this reality, this work focuses on drones or UAVs (Unmanned Aerial Vehicles) for use in industry in general. These vehicles have a large number of uses and potential in the industry, as a tool for civil engineering, medicine, mining, among others. However, this vehicle is limited for use indoors due to the need for GPS and it does not work indoors. In this way, this work presents a UAV that works without GPS, thus being able to be used in closed spaces for example and have good precision. The work is based on an approach that uses computer vision and GPS.

## 1 - INTRODUCCIÓN

La imitación del cerebro humano mediante algoritmos es una tecnología que tiene como objetivo que las máquinas puedan tomar decisiones propias sin que el humano lo manipule. Un sistema tenga la habilidad de presentar las mismas capacidades que una persona, ya sea movimientos, creatividad, razonamiento o poder de decisión, proporciona la oportunidad de utilizarlos en situaciones donde se necesita precisión o en tareas de alto riesgo para la persona.

Los vehículos aéreos no tripulados (UAV), popularmente conocidos como drones, forman parte de una gran área de innovación, integrados con cierta cantidad y variedad de sensores que permiten su uso en diversos sectores industriales y científicos. La inteligencia artificial y visión computacional dirigida a drones es un campo en desarrollo, un claro ejemplo es el proyecto AirSim de Microsoft "una nueva plataforma que se ejecuta en Microsoft Azure para construir, entrenar y probar de forma segura aeronaves autónomas a través de simulación de alta fidelidad." (Microsoft, 2022).

Como toda máquina, un dron también presenta riesgos en su manipulación, el choque con objetos o seres vivos es un evento que se debe evitar, problemas con la posición del vehículo ya está prácticamente estudiado. La utilización de GPS (Global Positioning System) y IMU (Inertial Measurement Unit) son tradicionalmente utilizados, proporcionando la información para poder controlar la trayectoria del dron. Sin embargo, esta manera presenta algunas debilidades si el GPS tiene un error o cuando hay demanda para uso *indoor*, por lo que esta incertidumbre puede generar un gran impacto en el uso que se le esté dando al dron.

Este proyecto presenta la aplicación de visión por computadora e inteligencia artificial dirigida a drones sin GPS. Buscamos estudiar cómo y qué herramientas utilizar para formar un dron con estas características, también cómo implementar la visión por computadora y la inteligencia artificial de forma que pueda desplazarse por un área, de forma que pueda modificar y recalcular su trayectoria según los objetos que se presenten en su camino. El vehículo propuesto tiene muchas posibilidades de uso en aplicaciones industriales, como por ejemplo ingresar a lugares desconocidos o espacios reducidos que presentan posibles peligros para el humano en donde hay diversos factores imprevisibles para el observador.

## 2 - METODOLOGÍA

Este proyecto incluye la construcción desde cero de un dron, este tiene como objetivo inicial la participación de competencias tanto nacionales como internacionales, pudiendo ser usado como apoyo a largo plazo en tareas *indoor* en la industria y comercio en general. Su elaboración fue llevada a cabo por integrantes de un equipo de competición en robótica, trabajando en el área de electrónica, programación y construcción. El laboratorio de este equipo está ubicado en la Universidad Tecnológica de Uruguay, Instituto Tecnológico Regional Norte (UTEC- ITR Norte), en la ciudad de Rivera, Uruguay. Allí se cuenta con un espacio adaptado para realizar tareas en el campo de investigación y construcción de robots con todas las herramientas al alcance.

Las principales competiciones de drones son Drone Racing League, Drone Champions League, Iberian Drone League y la Brazil Open Flying Robots League, la mayoría de estas instancias utilizan visión por computadora e inteligencia artificial dentro de sus categorías. En esas competencias, el desafío es enfocado en espacios reducidos donde existe la necesidad de evitar la colisión con objetos, eligiendo la mejor forma y camino para ello. El objetivo del equipo es participar en competencias donde no se nos evalúe a nosotros pilotando un dron ,sino a nuestra capacidad para obtener un resultado óptimo usando concepto de programación, IA y visión por computadora para el dron.

Más allá de nuestro objetivo de competición, el desarrollo de este dron es una gran oportunidad para investigar tanto en áreas de inteligencia artificial como en armado y manejo de drones, la idea es hacer el sim-to-real para el dron. Esto es, a partir de la simulación hacer las pruebas en un entorno real y así la comparación de los resultados evaluando desempeño. La participación de estudiantes y docentes del entorno, las oportunidades que brinda la frontera seca Santana do Livramento - Rivera, la ayuda de universidades más antiguas, entre otros factores, ayudan a lograr el enriquecimiento del conocimiento colectivo en la región. El resultado de este conjunto de colaboraciones y espacio de desarrollo de ideas será el crecimiento personal y profesional de cada participante, además de generar una red de conocimiento abriendo las puertas para una innovación exponencial en el área de robótica e inteligencia artificial.

**2.1 - DRONES**

La tecnología presente en los vehículos aéreos no tripulados abarcan su estabilidad, velocidad de vuelo, autonomía del mismo, telemetría, software, circuitos eléctricos, funcionalidad de sensores, entre otras cosas. Los UAVs tienen diferentes tamaños y funciones, también se pueden categorizar por cantidad de motores o por tipo de ala. Están los drones de ala fija, estos poseen un diseño de ala que permite al vehículo generar fuerzas sustentadoras para mantenerse en el aire, es decir que aprovechando el movimiento y la viscosidad del aire, logra una autonomía dada su eficiencia aerodinámica. Otros drones son de ala rotatoria, estos son los drones con los que trabaja el equipo de robótica de UTEC - ITR Norte, también conocidos como multirrotores, son los más utilizados por profesionales; estos generan la sustentación a través de la fuerza de los motores y las hélices. En esta categoría hay una separación según su número de rotores, tricópteros (3 motores), cuadricopteros (4 motores) como se muestra en la figura 1, etc.

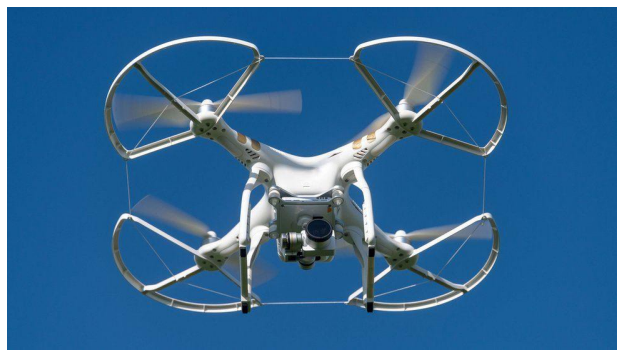

Figura 1. Dron cuadricoptero. (Kennisgeving voor omleiding, s. f.)

Se necesita de muchos componentes para poder volar un dron, esto es una desventaja debido al peso que suma cada una de sus piezas, cuanto más peso tenga el dron, más costoso es hacer que vuele, esto tiene una influencia directa con el consumo de energía y capacidad de carga, ya que cuantas más baterías integran el UAV mayor es el peso, a la hora de elaborar un dron se busca el menor peso posible.

**2.2 - HARDWARE**

El frame es el cuerpo del dron, teniendo como objetivo únicamente la construcción de un dron sin GPS y cámara para su uso en espacios reducidos y/o cerrados, el frame puede ser uno estándar como el de la figura 2. Sin embargo, si el uso que se le dará incluye presencia de personas, seres vivos u objetos que pueden dañar el dron, el diseño debe ser pensado de forma que no pueda lastimar nada o nadie y que al chocar no dañe el UAV (figura 3).

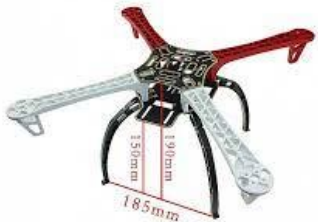

Figura 1. Frame F450 standard (Quadcopter F450 Frame with Landing Gear, 2021)

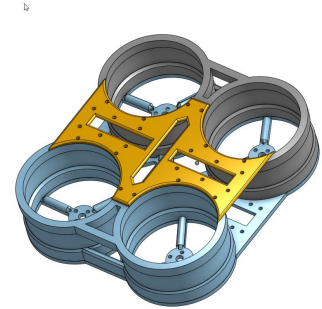

Figura 2. Frame diseñado por el equipo

PCB (Printed Circuit Board) o en español, Circuito Impreso, es un componente que ayuda a mantener compacto el diseño final del drone optimizando espacio. Este elemento es una tarjeta que soporta y conecta componentes electrónicos, utilizando caminos de material conductor (generalmente cobre).

Los ESC (Electronic Speed Control), es un circuito electrónico que permite controlar y regular la velocidad de motor conectado al mismo, cada motor tiene un ESC que regula la frecuencia del transistor para que la velocidad del motor cambie. Estos circuitos contienen un micro controlador que interpreta la señal de entrada y la traduce para controlar el motor. Figura 4.

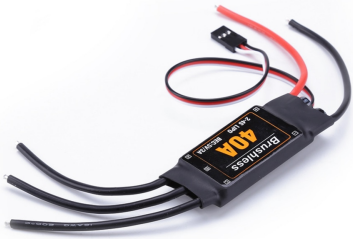

Figura 4. ESC (Variadores de velocidad ESC, s. f.)

La pixhawk es la placa controladora de los motores y sensores del dron, es un proyecto independiente de open-hardware, creado con el objetivo de su uso a investigación, pasatiempos o industrial, haciendo de este un producto low-cost que tiene una comunidad abierta a compartir información sobre su funcionamiento. Utilizando la pixhawk se consigue modificar los parámetros de vuelo del dron, corregir la estabilidad del dron y evaluar que esté funcionando todo correctamente para poder volar. Figura 5 muestra la placa controladora pixhawk y los componentes que van conectados a ella, no cual en nuestro caso hay destaca para la no utilización de GPS, controlador o pantalla.

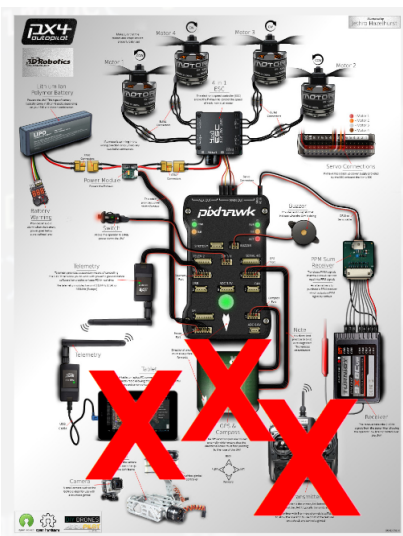

Figura 5. (Advanced Pixhawk Quadcopter Wiring Chart — Copter documentation, s. f.)

Cámara ZED, es una cámara 3D de alto rendimiento que permite ubicar al dron en el espacio y ver la estabilidad del dron durante su vuelo. Mediante software es posible utilizar funciones como detección y reconocimiento de objetos, tiene muchas aplicaciones, nosotros la utilizaremos para lo antes nombrado.

Jetson Nano, es una pequeña y potente computadora, desarrollada con el objetivo de implementar sistemas de inteligencia artificial. Nuestro dron es la computadora embarcada que realiza todo el procesamiento del software de IA y visión computacional, calculando la odometría visual que es enviada a la PixHawk. (¿Qué puedo hacer con una Jetson Nano?, 2020b)

**2.3 - SOFTWARE**

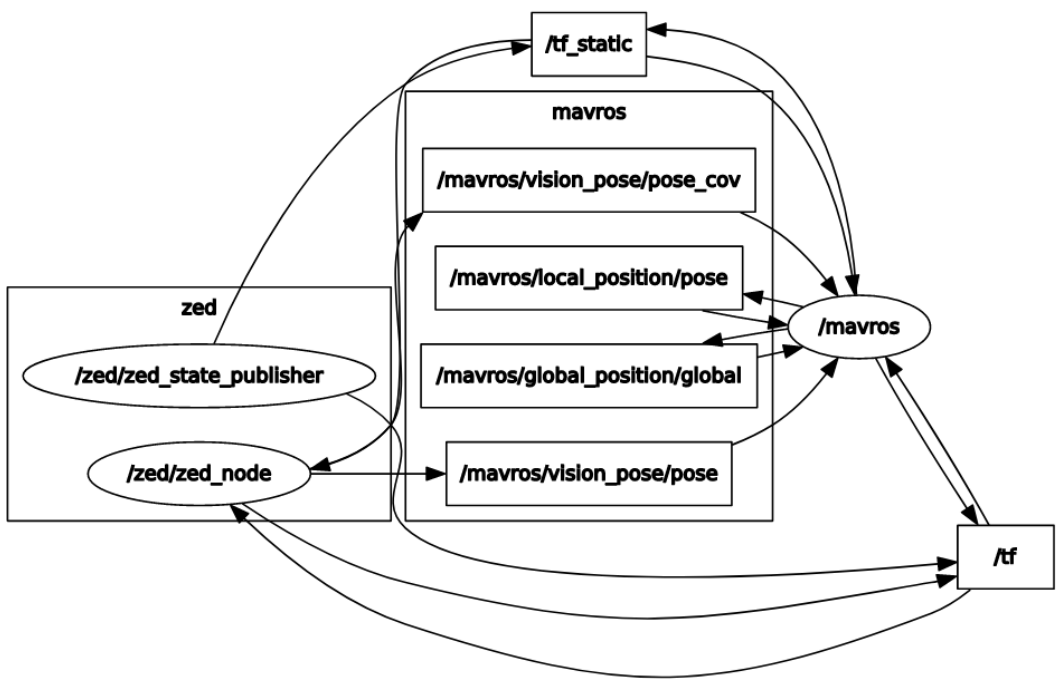

Figura 6. Esquemático del Software del Dron que fue utilizado.

El software del dron fue desarrollado utilizando el Sistema Operacional de Robots (ROS). El ROS es un meta sistema operacional que funciona en Linux y tiene soporte a diversas placas y circuitos integrados, como por ejemplo la PixHawk y la Jetson. Además, ROS también permite hacer códigos en distintos lenguajes como por ejemplo Python y C++. Para nuestro Dron, fue utilizado ROS Noetic con código en python para hacer la Inteligencia de navegación y C++ para hacer la odometría visual con la cámara ZED.

**3 - TESTS PRÁCTICOS**

El dron desarrollado fue utilizado para lograr el tercer lugar en el Hackathon de Drones Uruguay 2022, que ocurrió en Mercedes - Uruguay - en el mes de octubre. Figura 7 muestra el equipo Urubots en la competencia y en el trabajo se adjunta un video (https://youtube.com/shorts/j_Hn7kW4kfU?feature=share) mostrando el test de vuelo con el drone totalmente autónomo y con odometría visual.

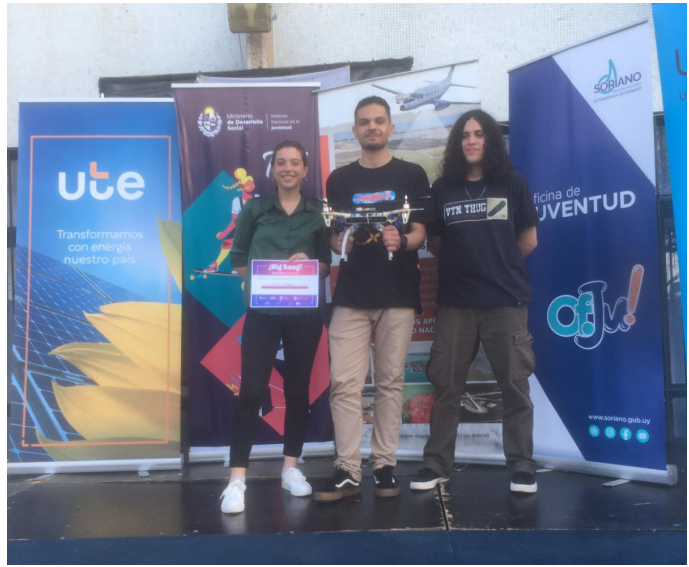

Figura 7. El dron desarrollado fue usado para lograr el tercer lugar en el Hackathon de Drones Uruguay 2022.

**4 - CONCLUSIÓN**

Luego de haber trabajado en este proyecto durante los últimos meses, ha sido grande el avance tanto en área de conocimiento para los integrantes del equipo como para el entorno académico en el que se encuentra. Este dron permanece en construcción más allá de haber hecho las primeras pruebas donde se logró que este vuele en un frame standard, todavía faltan recursos y adaptaciones para lograr una versión que cumpla con los objetivos planteados.

Fue posible desactivar el GPS integrado en la pixhawk, combinado con 4 ESC funcionales e iguales para cada uno de los motores, habiendo diseñado el software el cual ejecutar en la Jetson Nano, se logró que el dron además de volar, logre mantenerse en el aire corrigiendo su vuelo y estabilidad gracias a la cámara ZED.

Estos avances significativos impulsan la creatividad del equipo a buscar aplicaciones y soluciones a errores que puedan surgir en el camino, buscando cada vez más involucrarse en el área de drones y especializándose en temas relacionados. Mostrando el compromiso y logros en relación a este proyecto, será posible atrapar el interés de empresas y organismos con posibilidad de invertir, pudiendo así ayudarnos a enriquecer el conocimiento no solo sobre drones sino en robótica e inteligencia artificial.

**BIBLIOGRAFÍA**